\documentclass[runningheads]{llncs}
\usepackage[T1]{fontenc}
\usepackage{graphicx}
\usepackage{booktabs}
\usepackage[ruled,vlined]{algorithm2e}
\usepackage{amsmath}
\usepackage{amsfonts}

\newcommand{\E}{\mathbb{E}}
\DeclareMathOperator*{\argmax}{arg\,max}
\newcommand{\pr}{\mathbb{P}}

\begin{document}
\title{A Policy for Early Sequence Classification}
%
%
\author{Alexander Cao\inst{1} \and
Jean Utke\inst{2}  \and
Diego Klabjan\inst{1}}
\authorrunning{A. Cao et al.}
%
\institute{Department of Industrial Engineering and Management Sciences, Northwestern University, Evanston IL, USA \\
\email{a-cao@u.northwestern.edu, d-klabjan@northwestern.edu} \and
Data, Discovery and Decision Science, Allstate Insurance Company, Northbrook IL, USA \\
\email{jutke@allstate.com}
}
\maketitle              
\setcounter{footnote}{0}
\begin{abstract}
Sequences are often not received in their entirety at once, but instead, received incrementally over time, element by element. Early predictions yielding a higher benefit, one aims to classify a sequence as accurately as possible, as soon as possible, without having to wait for the last element. For this early sequence classification, we introduce our novel classifier-induced stopping. While previous methods depend on exploration during training to learn when to stop and classify, ours is a more direct, supervised approach. Our classifier-induced stopping achieves an average Pareto frontier AUC increase of 11.8\% over multiple experiments.

\end{abstract}
\section{Introduction}
Practical use cases for early sequence classification exist in many domains. Holding your smartphone’s microphone up to a speaker, in seconds a music recognition app can tell which song is being played. There are two competing objectives with respect to the app making a real-time classification from audio. On one hand, a longer sequence from the song may yield a more accurate classification. On the other hand, the user may not have the patience to wait very long. 

Generally, we are interested in scenarios in which a classifier receives elements of a sequence over time. This kind of ongoing flow of data immediately suggests a need for a real-time ability to stop waiting for new elements and classify given the received elements at this point in time at sufficient accuracy. We call this early classifying to differentiate from classification after a `complete' sequence or a pre-set number of sequence elements is received. Optimally deciding when one has received enough data, and then making an accurate classification from that data, is the crux of the problem we are investigating. 

To this end, we introduce our novel classifier-induced stopping (CIS) in this paper. Previous methods depend on exploration during training (when there is access to the entire sequence) to learn (i) a policy to decide when to stop waiting for new elements and classify and (ii) the classifier itself. Exploration, in an early sequence classification context, means the policy affects how much of the sequence is ingested or used to learn. In contrast, CIS learns both policy and classifier in a more direct, supervised approach inspired by imitation learning \cite{attia2018global}. CIS learns to classify as accurately as possible at every time step, after receiving a new element. Concurrently, it learns to stop and classify at the optimal time (based off a reward) induced from its own classifications at each time step. CIS removes notions of exploration and learns to follow the ideal decision-making based off its own classification predictions; hence, we call it classifier-induced. The main contributions of our work are as follows. We introduce a novel, supervised framework to learn a stopping time for early classifiers that avoids exploration. Instead, it learns when to stop from its own classifications. We demonstrate that CIS outperforms benchmarks in terms of a Pareto frontier AUC measure across diverse experiments. 

Our paper is structured as follows. In Section \ref{sec:relatedWork}, we establish notation and review related work, specifically the two benchmark methods used in experiments. Following in Section \ref{sec:cis}, we discuss CIS in detail. Section \ref{sec:exp} presents results from three sets of experiments on a variety of problems and data. Section \ref{sec:conclude} gives a summary.

\section{Related Work} \label{sec:relatedWork}
\subsection{Problem Setup Notation} \label{sec:notation}
The framework of early classification we consider here is as follows. The set of training data $\mathcal{X}$ comprises sequences $x^{(i)}$ paired with one-hot encoded labels $y^{(i)} \in \{0,1\}^{C}$ for $C$ classes, where $x^{(i)}= \left(x^{(i)}_1,  x^{(i)}_2, ..., x^{(i)}_{T_\text{end}} \right)$ is a sequence of tensors. At time $t\leq T_\text{end}$, its state is given by $s^{(i)}_t = \left(x^{(i)}_1,  x^{(i)}_2, ..., x^{(i)}_{t} \right)$.

A {\em classifier} neural network $  f_\alpha \left( s_t \right) = \widehat{y}_\alpha \left( \cdot | s_t \right) $ parametrized by $\alpha$ takes $s_t$ as input\footnote{We omit the $^{(i)}$ indices unless needed.} and outputs predicted class distribution vector $ \widehat{y}_\alpha \left( \cdot | s_t \right)$ at time $t$. 
A {\em policy} neural network $g_\beta \left( s_t \right)  = \pi_\beta \left( \cdot | s_t \right)$ parametrized by $\beta$ takes $s_t$ as input and outputs policy distribution vector $\pi_\beta \left( \cdot | s_t \right) $ over two actions (`wait' and `stop and classify') at time $t$. 

At each time step $t$, we take an action $a_t$ according to policy $\pi_\beta \left( \cdot | s_t \right)$. This is done stochastically via sampling or deterministically via taking the most likely action. We keep waiting another time step and receive new element $x_{t+1}$ until we decide to stop. Once we decide to stop and classify, we make a classification according to $ \widehat{y}_\alpha  \left(\cdot | s_t \right)$.  To encourage a model to early classify as accurately as possible, as quickly as possible, we use the following reward function at each time step $t$
\begin{align}
\begin{split}
&R_t^\alpha (s_t, a_t) \\
&= \begin{cases}
-\mu \quad \text{if $a_t = \text{`wait'}$} \\
-\mu - \text{CE} \left(y,   \widehat{y}_\alpha \left(\cdot  | s_t \right)  \right) \quad \text{if $a_t = \text{`stop and classify'}$ or $t=T_\text{end}$} 
\end{cases}
\end{split}
\end{align}
where $\mu$ is a time penalty parameter and CE is cross-entropy. At each time step, a constant penalty of $-\mu$ is incurred. Early classification is completed once the model decides to stop and classify at a time $T$. The problem is to solve
\begin{align}
\max_{\alpha, \beta} \E_\mathcal{X} \sum_{t} R_t^\alpha \left( s_t, a_t \left( \beta \right) \right). \label{eq:cumulativeReward}
\end{align}
Maximizing the cumulative reward is equivalent to classifying as accurately as possible (so that the cross entropy is low), as quickly as possible (so that the sum of time penalties is low). The time penalty parameter $\mu$ controls how much waiting another time step is penalized. If $\mu$ is large, we may sacrifice more accuracy for an earlier classification, and vice-versa.  

The problem has two challenges. When a policy decides to stop, it never directly learns what would happen if it waited longer. In essence, the ability to look forward and learn from information after the stopping time is important. Second, and more subtly, the policy and classifier need to be cohesively learned together as the time penalty relates the two. 

\subsection{Early Classification via Reinforcement Learning}
Several papers treat early classification as a standard reinforcement learning problem. \cite{liu2020finding} ingests text sentence-by-sentence and answers given questions (via classification) when the model decides enough information has been read. \cite{hartvigsen2019adaptive} applies a very similar methodology to obtain early diagnoses from healthcare vital signs like EEGs. It is important to note that \cite{liu2020finding,hartvigsen2019adaptive} still train their models with the REINFORCE algorithm \cite{williams1992simple}, a standard policy gradient method. They compare against full-sequence-length classifiers or utilize a fixed threshold on each time step's classification as a stopping rule. We choose the Proximal Policy Optimization (PPO) algorithm \cite{schulman2017proximal} as our standard reinforcement learning benchmark to compare against CIS; details are in Section \ref{sec:ppo}.

\subsection{PPO} \label{sec:ppo}
Policy gradient methods work by first creating episodes 
$$(s_1, a_1, R^{\alpha}_1), (s_2, a_2, R^{\alpha}_2), ..., (s_T, a_T, R^{\alpha}_T)$$ 
with actions determined by the current policy. The policy is then updated in gradient ascent direction so that actions leading to greater future rewards become more probable. PPO, following \cite{schulman2017proximal}, maximizes the clipped surrogate objective
\begin{align}
\begin{split}
&\mathcal{L}_\text{PPO} \\
&=  \E_{\mathcal{X}, t} \left[ \min\left\{ \frac{\pi_\beta \left( a_t | s_t \right) }{\pi_{\beta_\text{old}} \left( a_t | s_t \right) } \widehat{A}_t^\alpha, \text{clip} \left( \frac{\pi_\beta \left( a_t | s_t \right) }{\pi_{\beta_\text{old}} \left( a_t | s_t \right) }, 1-\epsilon, 1+\epsilon\right)  \widehat{A}_t^\alpha\right\} \right].
\end{split}
\end{align}
The estimated advantage $\widehat{A}_t^\alpha$ is given by $\widehat{A}_t^\alpha = \sum_{t'=t}^T \gamma^{t'-t} R_{t'}^\alpha - V\left(s_t \right)$ where $\gamma$ is a discount factor and $V\left(s_t \right)$ is a learned state-value function. PPO's exploration hindrance is evident as any information after time $T$ is not used in learning. Keeping in line with previous work relying on exploration, we opt to keep the policy stochastic during inference \cite{hartvigsen2019adaptive,schulman2017proximal,huang2017length}. 

Policy gradient reinforcement learning methods are, by nature, trial and error-based. They cannot take advantage of the fact that stopping and classifying later for a given sample would have been better. Put differently, they do not utilize the entire sequence during training. 

\subsection{LARM}
Length Adaptive Recurrent Model (LARM) \cite{huang2017length} and CIS remedy this inability to look forward in the sequence. LARM takes a more probabilistic interpretation to early classification. Let $A_T = \left(a_1 = \text{`wait'}, a_2=\text{`wait'}, ..., a_{T-1} =\text{`wait'}, \right.$ \\
$\left. a_T=\text{`stop and classify'} \right)$ be a decision sequence where the policy decided to wait the first $T-1$ time steps and stopped to classify at time $T$. Given $A_T$ and $\pi_{\beta} \left( \cdot | s_t \right)$, we can explicitly factor the probability of sequence $A_T$ as
\begin{align}
\pr \left( A_T | s_T \right) = \prod_{t=1}^{T} \pi_{\beta} \left( a_t | s_t \right).
\end{align}
With respect to this stopping time probability, LARM seeks to maximize the expected cumulative reward in \eqref{eq:cumulativeReward} with the objective 
$$\max_{\alpha, \beta} \E_\mathcal{X} \left[ - \text{CE} \left(y,  \sum_{T=1}^{T_\text{end}} \left( \widehat{y}_\alpha | s_T \right) \pr \left( A_T | s_T \right)  \right) - \mu  \sum_{T=1}^{T_\text{end}} T \cdot \pr \left( A_T | s_T \right) \right].$$ 
The first term is a micro-averaged cross-entropy loss and the second term is the expected stopping time. 

Because $\pr \left( A_T | s_T \right)$ is a product whose value may exponentially decrease, LARM takes special care to prevent this. During training, the factors \\
$\pi_{\beta} \left( a_t  = \text{`wait'} | s_t \right)$ are set to 1 with probability $\rho$. This forces the model to wait for more elements in the sequence and not get stuck stopping too soon. In terms of early classification, waiting is tantamount to ingesting more information and so $\rho$ is a parameter controlling this aspect. Even so, there is an exploration drawback here in that learning accurate classifications at low probability stopping times is difficult. For inference, LARM opts for stochastic policy rollout with deterministic classification.

\section{Classifier-Induced Stopping} \label{sec:cis}
\begin{figure}[h!]
\centerline{
\includegraphics[width=0.5\linewidth]{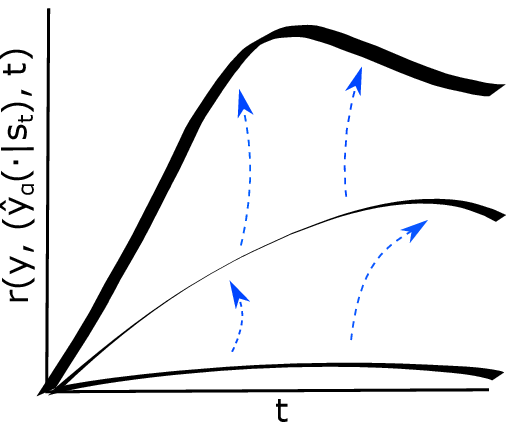} 
\includegraphics[width=0.5\linewidth]{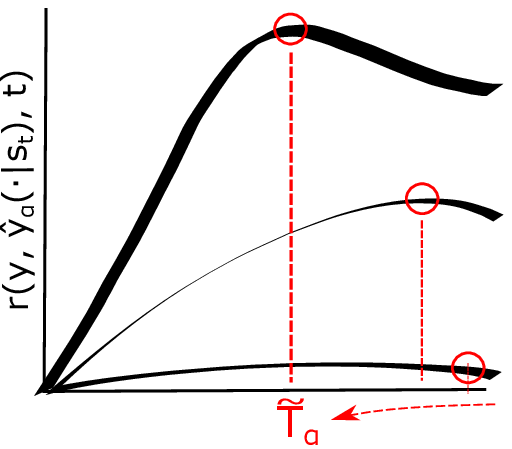} 
}
\caption{(Left) $\mathcal{L}_{\widehat{y}}$ is increasing cumulative reward $r$ at each time step. (Right) Concurrently, for the rendered reward curve, there exists an optimal time to stop and classify $\widetilde{T}_\alpha$ that maximizes $r$ and therefore an optimal policy $ \widetilde{\pi}_\alpha \left( \cdot | x,t \right)$. $\mathcal{L}_{\pi}$ aims to learn this policy.}
\label{fig:L_CIS}
\end{figure}

As previously stated, early classification can be framed as maximizing the cumulative reward given in \eqref{eq:cumulativeReward}. We can recast this quantity as a function $r$ depending on label $y$, classification prediction $\widehat{y}_\alpha   \left(\cdot | s_T \right) $, and classification time $T$ given by  $r\left( y, \widehat{y}, T \right) = -\text{CE}\left( y, \widehat{y} \right) -\mu T$. Note, for a fixed $\widehat{y}$ and $y$ this is a univariate function of time $T$. With this in mind, we aim to learn (i) when to stop and classify and (ii) what classification to make in a more direct, supervised manner. First, CIS seeks to make the most accurate classification prediction at every single time step. Second and simultaneously, CIS learns the corresponding policy which yields the resulting optimal classification time. In this way, our policy learns the ideal policy based off of its own classifications. Hence, we name it classifier-induced. The loss function is given by $\min_{\alpha, \beta} \mathcal{L}_\text{CIS}  = \min_{\alpha, \beta} \E_\mathcal{X} \left[  \mathcal{L}_{\widehat{y}} +  \lambda  \cdot  \mathcal{L}_{\pi}  \right]$ where
\begin{align}
\mathcal{L}_{\widehat{y}}  = \frac{1}{T_\text{end}} \sum_{t=1}^{T_\text{end}} \text{CE}\left( y,   \widehat{y}_\alpha \left( \cdot | s_{t} \right) \right)  \label{eq:Lyhat}, \quad \mathcal{L}_{\pi} = \frac{1}{T_\text{end}} \sum_{t=1}^{T_\text{end}} \text{CE}\left( \widetilde{\pi}_\alpha \left( \cdot | x, t  \right) , \pi_\beta \left( \cdot | s_t \right) \right)   \\
\widetilde{\pi}_\alpha \left( a |x,t \right)  = \begin{cases}
(1, 0) \quad \text{if $t<\widetilde{T}_\alpha \left(x\right)$}\\
(0,1) \quad \text{if $t \geq \widetilde{T}_\alpha \left(x \right)$},
\end{cases} \widetilde{T}_\alpha \left(x \right) = \argmax_t r\left( y, \widehat{y}_\alpha  \left(\cdot | s_{t} \right), t  \right) .
\end{align}
We write $(1,0)$ to mean `wait' with probability 1 and $(0,1)$ as `stop and classify' with probability 1. There is hyperparameter $\lambda$. Figure \ref{fig:L_CIS} below offers an intuitive visual walkthrough of CIS.

Unlike PPO and LARM, our novel CIS does not rely on any notion of exploration. The entire sequence is wholly used in training and we are able to directly learn the optimal classification time in a supervised manner. During training $ \widetilde{\pi}_\alpha \left( \cdot | x,t \right)$ and $\widetilde{T}_\alpha$ are treated as fixed labels in minibatch updates. Since there is no exploration in CIS, the policy does not have an exploratory nature; hence, in inference we simply take the argmax action.

\section{Experimental Results} \label{sec:exp}

\subsection{Datasets and Pareto Metric}
Our first experiment is with the IMDB movie reviews sentiment analysis dataset \cite{maas-EtAl:2011:ACL-HLT2011}. We do not need to ingest the entire review to classify its sentiment. Instead, we read word by word and classify the review after ingesting a minima number of words. The dataset comprises 50,000 movie reviews; half the reviews are positive and the other half negative. We reserve a random 15\% of samples to be the hold-out validation set, separate from the training set. We set $T_\text{end} = 236$, which is the mean training review length, and pad up or truncate down all reviews to this length. 

The second experiment uses Electrocardiography (ECG) waveforms of multiple cardiovascular diagnoses from PTB-XL \cite{wagner2020ptb}. ECGs record electrical signals from the heart and help to assess cardiac clinical status of patients. Instead of a diagnostic tool alone, early classification aids in continuous monitoring for heart conditions. The sooner an early classifier can detect a heart attack, the sooner medical attention can be given. Here we early classify ECG signals by ingesting small segments sequentially. After following \cite{smigiel2021ecg} and filtering out some ECGs (those with uncertain diagnoses, for instance), we are left with 17,221 samples in the dataset. There are five classes which are reasonably balanced. We reserve a random 10\% of samples to be the validation set, again separate from the training set. Each ECG length is 10 seconds, sampled at 100 Hz. Consistent with the procedure in \cite{zihlmann2017convolutional}, the network input is the log spectrogram of each ECG (using a Tukey window of length 32 with 50\% overlap). In essence, spectrograms are consecutive fragments of a signal in Fourier space to represent frequencies varying over time. Therefore our early classifier, in effect, receives each ECG in consecutive 0.16 second fragments in Fourier space.

Our third and final experiment is motivated by European call options. They give the holder the right to buy a stock at a specified strike price only on a given expiration date (betting the stock will go up). However, after buying the option, if the option holder could predict that the stock price will not be above the strike price on the option expiration date, then the holder could attempt to sell the option in the secondary market to recoup the original cost of the option. To be clear, in this problem context we are concerned only with the prediction aspect and not the option sale.

From \cite{huang2004specification}, it is reasonable to assume a strike price equal to the stock price on the option origination purchase date. With this in mind, we simulate 1-month European call options in the following way. Samples are generated from 65 current S\&P 500 technology stocks based on daily data ranging from 1962 to 2017. For the training set, we divide each technology stock into disjoint 30-day stock price samples, through 2016. We consider a binary classification of whether the stock closing price on day 30 is greater than or less than the stock closing price on day 1 (proxying strike price). Thus, stopping to classify is akin to committing on day $T$ to exercise the option or not upon expiration. This process yields 9,313 training samples with 59\% of these options as profitable to exercise. For validation, we wish to roll out the early classifier more organically and continuously. Accordingly, we take the remaining year 2017 after the training set from each stock for validation. The assumption is that we will have year-long stock price sequence to continually roll out early classifiers and `purchase' new options the day after stopping and deciding what to do with the current one. Table \ref{tab:stockFeatures} summarizes daily technical indicators used along with the standard open, high, low, and close prices plus volume to form the daily features.

\begin{table}[h!]
  \caption{Stock price sequence technical indicators, using standard parameters}
  \label{tab:stockFeatures}
  \centering
  \begin{tabular}{l l}
    \toprule
    Feature     & Description      \\
    \midrule
     Exponential moving average & Measures trend direction, \\
     \quad (open, high, low, close, volume) & \quad heavier weighting on more recent days \\
     Bollinger Bands & Relative highs and lows of price movement \\
     On-balance volume & Measures buying and selling of stock \\
     Accumulation/distribution & Gauges supply and demand  \\
     Average directional & Measures trend strength  \\
     Aroon oscillator & Indicates uptrend or downtrend  \\
     Moving average &  Measures momentum \\
     \quad convergence/divergence &  \\
     Relative strength & Measures speed of price changes  \\
     Stochastic oscillator & Measures momentum  \\
    \bottomrule
  \end{tabular}
\end{table}

In all experiments, we holistically compare early classifiers from PPO, LARM, and CIS by their Pareto frontiers. This allows us to examine the entire performance spectrum of their accuracy-timeliness tradeoffs. Our procedure for constructing a Pareto frontier is as follows. For a given $\mu$ value, we roll out the early classifier over the validation set and compute the mean classification time and accuracy after each training epoch. This is repeated for varying $\mu$ to get the entire collection of such accuracy-timeliness tradeoff points. Finally, all dominated points are removed which yields the Pareto frontier. The Pareto frontier (piecewise-constant) AUC is a holistic measure of accuracy-timeliness tradeoff efficacy. We treat $\mu$ as a hyperparameter, controlling the dichotomous balance between accuracy and timeliness, and sweep multiple values to trace the Pareto frontier. In a real-world use case, extrinsic factors from the problem itself should guide which Pareto point is optimal.

\subsection{Implementation}
Before describing network design and hyperparameters, we modify PPO's objective. Learning a state-value baseline function leads to more unstable training and ultimately poorer results. So in our case we remove it, and the advantage reduces to the sum of future rewards $\widehat{A}_t^\alpha = \sum_{t'=t}^T \gamma^{t'-t} R_{t'}^\alpha$. In addition to PPO's main objective, we add a classification term to help directly teach the classifier. The combined objective is then 
$$\min_{\alpha,\beta}\big( \E_{\mathcal{X}, t} \left[ \text{CE} \left(y,    \widehat{y}_\alpha \left( \cdot | s_t \right)  \right) \right] -\mathcal{L}_\text{PPO}\big)\quad.$$ 

While the policy and classifier can be disjoint networks, in practice it is common to have them as two heads of the same body network \cite{huang2017length}. We choose this for our implementations of PPO, LARM, and CIS, with the body network being an LSTM. Elements of sequential data (or embeddings) are inputs to the LSTM. The recurrent hidden states are in turn inputs to separate, feed-forward networks: one for the policy and one for the classifier. Each of these feed-forward heads is composed of a single hidden-layer with ReLU activation and softmax output. Next, we explicate all of the hyperparameters used in our experiments.

For all three experiments, we sweep $\mu \in \left\{ 0.001, 0.003, 0.005, 0.007, 0.01,   \right.$ $\left. 0.03, 0.05, 0.07, 0.1 \right\}$. We keep the standard PPO clip value of $\epsilon=0.2$ and a discount factor $\gamma=1$ yields the best results (and does not scale the cumulative reward). To further aid PPO waiting longer and ingesting more information initially, yielding better results, we initialize the policy head's final layer's bias to $(10,0)$. For CIS, we set the scaling constant $\lambda=1$. The training set is optimized by using Adam with batch size 128 until validation accuracies and mean classification times plateau. 

For the IMDB experiment specifically, the word embedding dimension is 32. For the network size, the LSTM hidden state is of dimension 64 and the two FFN hidden layers are of dimension 32. Learning rates for PPO and LARM are $10^{-4}$ and $10^{-3}$ for CIS. Following \cite{huang2017length}, we keep LARM's waiting parameter $\rho=0.9$.

For the ECG experiment specifically, the hidden vector of the LSTM is of dimension 128 and the two FFN hidden layers are of dimension 64. All three learning rates are set to $10^{-4}$. Again, we keep LARM's $\rho=0.9$. 

Finally, for the stock option experiment specifically, we implement a chronological rolling normalization so that all features are scaled in range $[0,1]$. The network dimensions are 32, 16, and 16. All three learning rates are set to $10^{-4}$. In this experiment, LARM performs poorly with $\rho=0.9$ and lowering it to $\rho=0.6$ lead to significantly better performance.\footnote{We pledge to publish our code and add a link here upon acceptance of this paper.} 

\subsection{IMDB Experiment}
\begin{figure}[h!]
\centerline{
\includegraphics[width=0.5\linewidth, trim=0.3cm 0.5cm 2.0cm 2.2cm, clip]{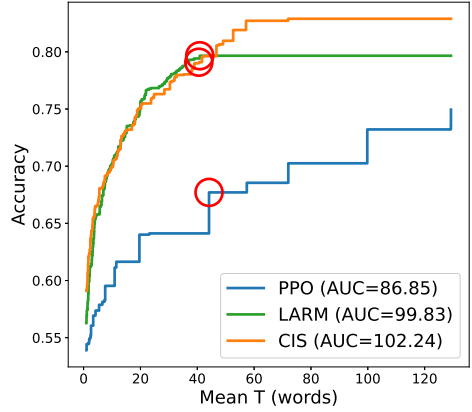} 
}
\caption{Pareto frontiers for the IMDB experiment.}
\label{fig:imdbPareto}
\end{figure}

Figure \ref{fig:imdbPareto} displays the Pareto frontiers for the IMDB experiment. CIS's AUC is 17.7\% greater than PPO's AUC and 2.4\% greater than LARM's AUC. CIS outperforms PPO and LARM, and we stress this due to the forward-looking, supervised nature of the algorithm. 

\begin{figure}[h!]
\centerline{
\includegraphics[width=0.5\linewidth, trim=0.4cm 2.7cm 1.7cm 2.2cm, clip]{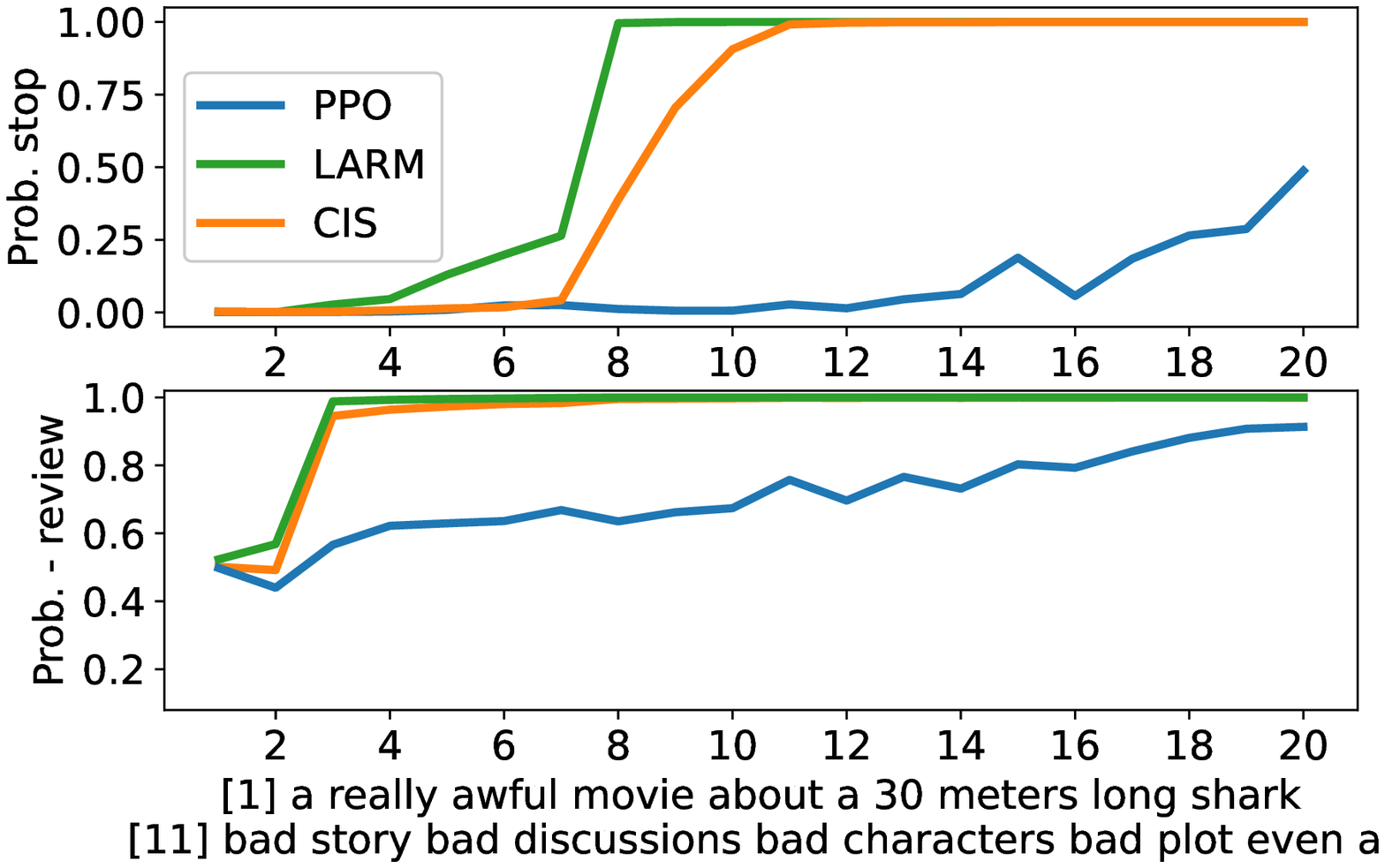} 
\includegraphics[width=0.5\linewidth, trim=0.4cm 2.7cm 1.7cm 2.2cm, clip]{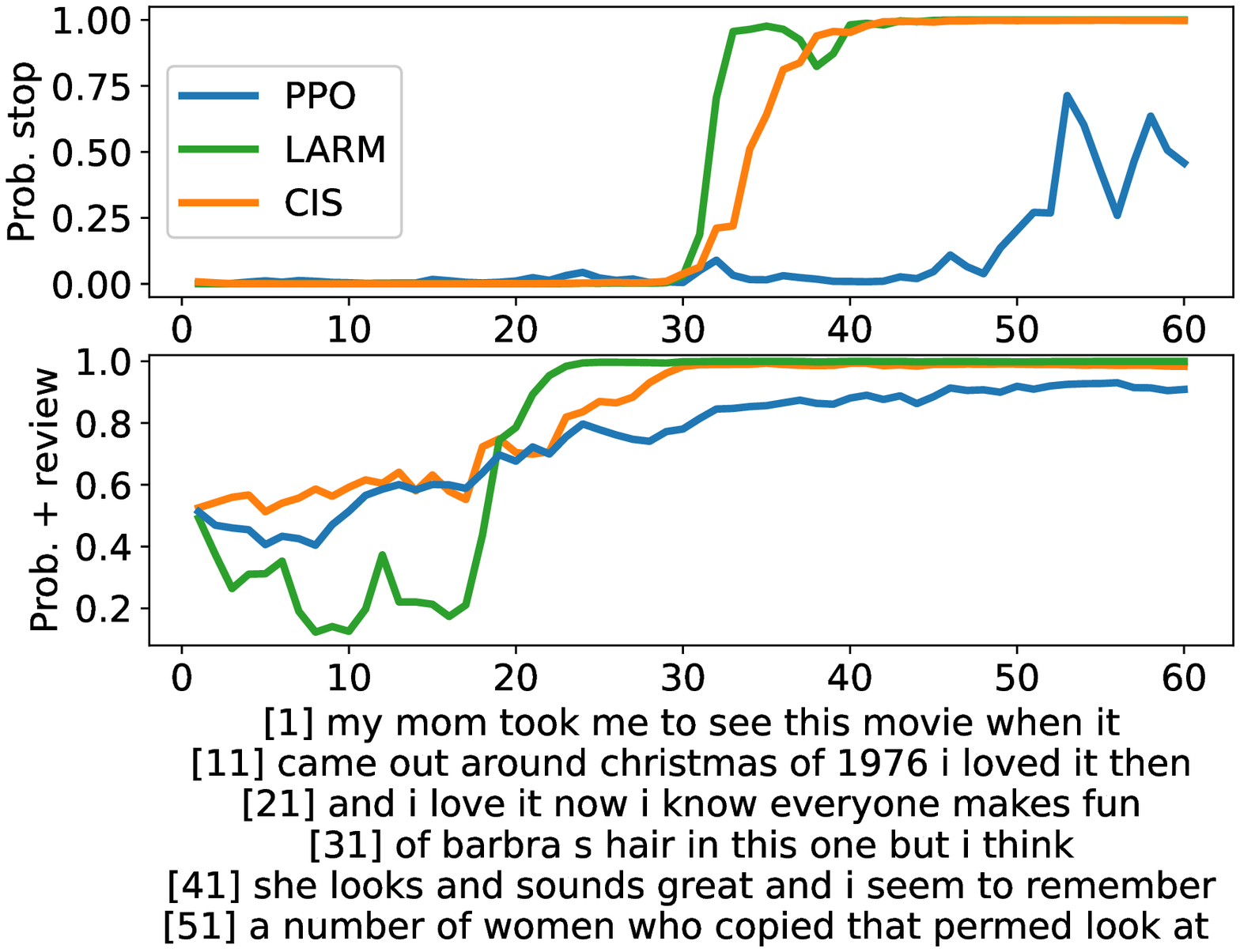} 
}
\caption{Early classifer performances on an example (left) stark negative and (right) long-winded positive IMDB movie review.}
\label{fig:imdbEx1}
\end{figure}

While CIS and LARM performs well globally, they are also coherent on an individual review level. Let us consider each early classifier at a mean $T$ of about 40 words (red circles in Figure \ref{fig:imdbPareto}). In Figure \ref{fig:imdbEx1} we present their respective outputs for a very stark negative movie review. For this review, CIS and LARM are able to quickly and correctly stop soon after `a really awful movie' while PPO continues to wait. Additionally, in Figure \ref{fig:imdbEx1} we also show CIS and LARM's abilities to early classify a long-winded, positive movie review. The first 20 words in this review are not actually about the movie itself. It is not until `i loved it then and i love it now' that the models sense the review's sentiment and act and classify accordingly. Again PPO seems to need more information. These two didactic examples indicate the discerning patience and linguistic understanding of CIS and LARM over PPO, contributing to the gap in accuracies.

\subsection{ECG Experiment}
\begin{figure}[h!]
\centerline{
\includegraphics[width=0.5\linewidth, trim=0.3cm 0.5cm 2cm 2.2cm, clip]{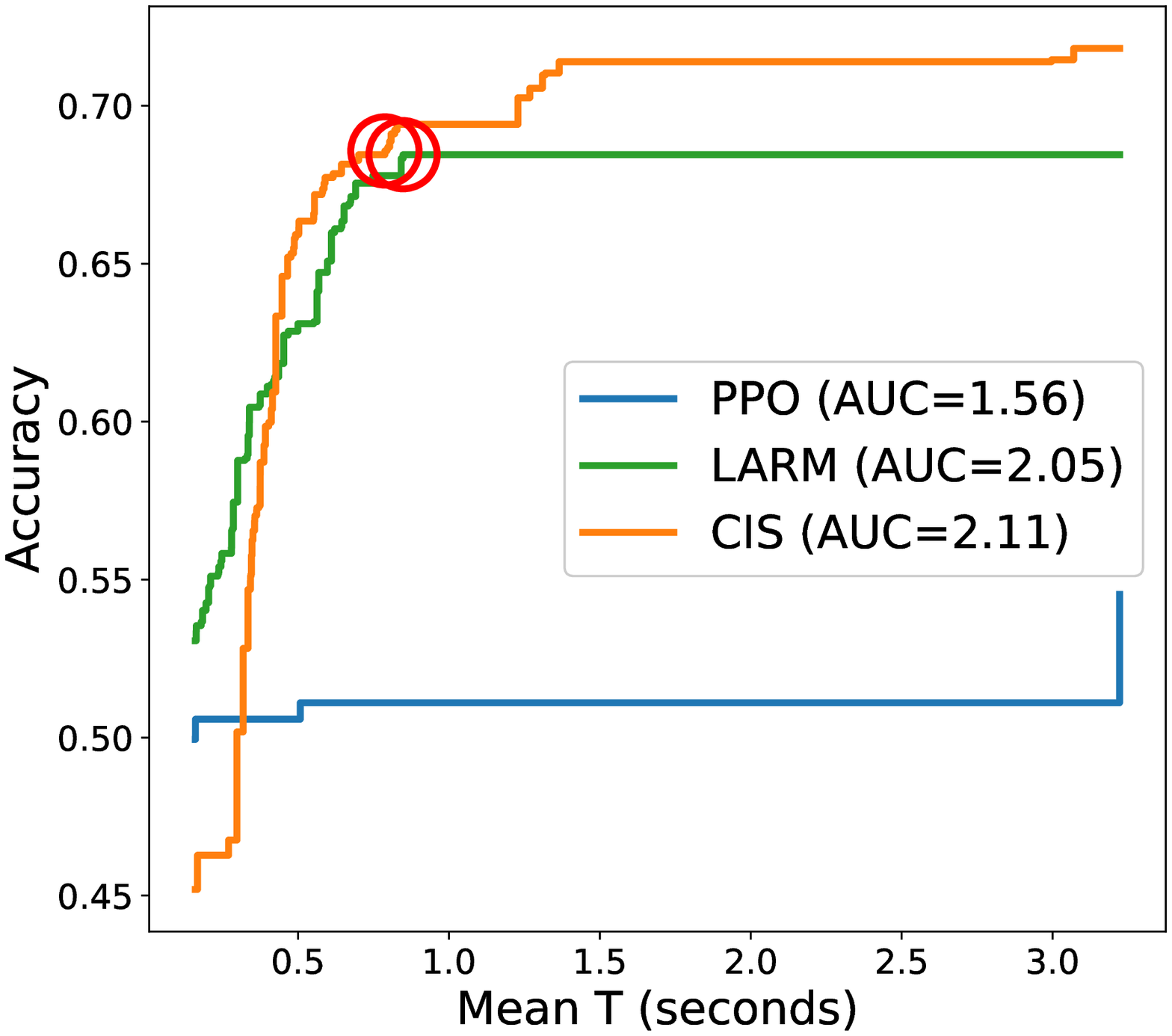} 
\includegraphics[width=0.5\linewidth, trim=0.6cm 0.4cm 2cm 2.2cm, clip]{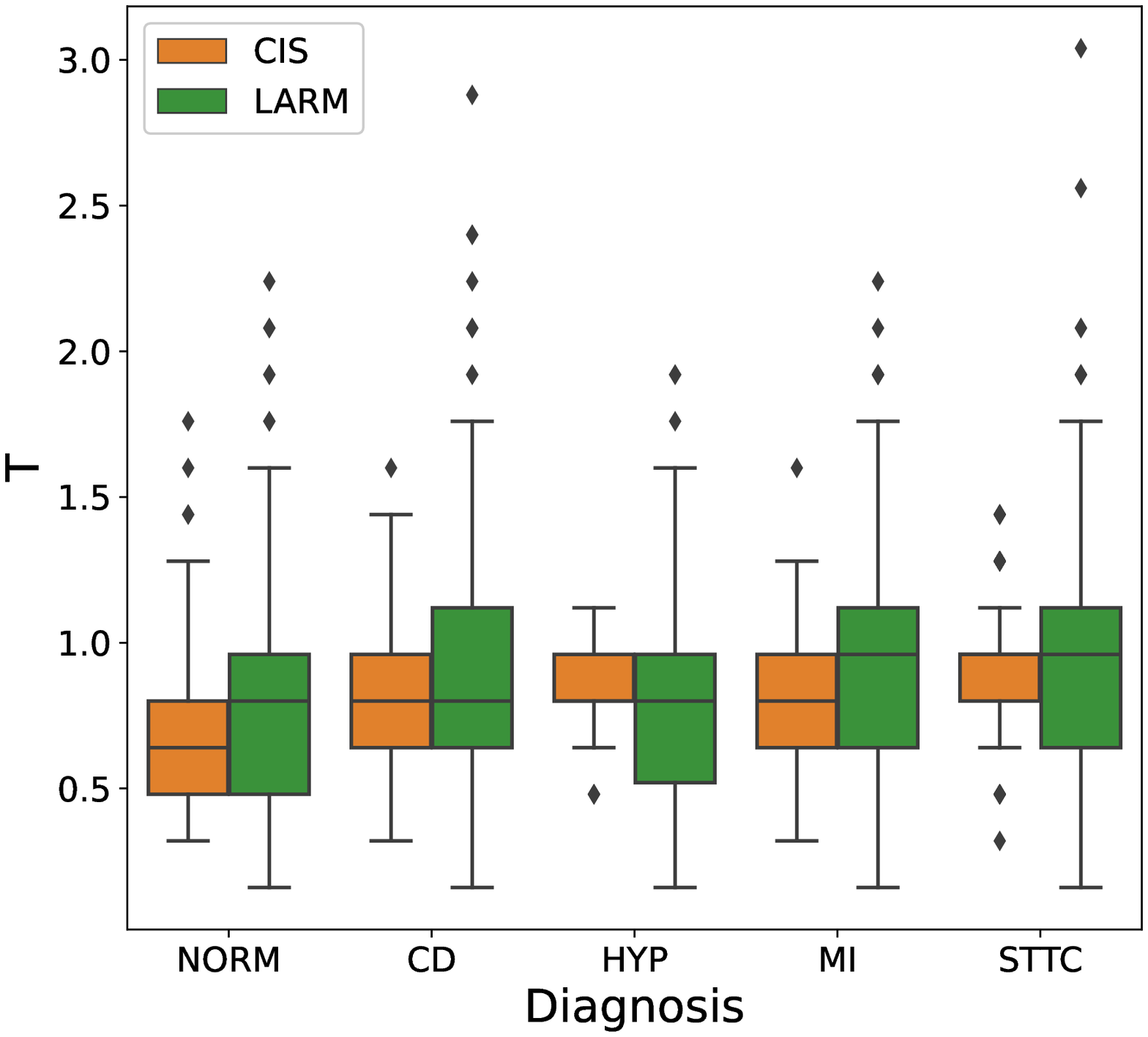} 
}
\caption{(Left) Pareto frontiers for the ECG experiment. (Right) Box plots showing distribution of CIS and LARM classification times $T$ for each ground truth ECG diagnosis. For HYP, CIS's first quartile coincides with the median because of repeating values. Similarly for STTC, CIS's third quartile coincides with the median.}
\label{fig:ecgPareto}
\end{figure}

Figure \ref{fig:ecgPareto} (left) displays the Pareto frontiers for the ECG experiment. CIS holistically outperforms PPO and LARM. CIS's AUC is 35.3\% greater than PPO's AUC and 2.9\% greater than LARM's AUC. Although it is worthwhile to note that CIS performs worst for mean $T$ below 0.3 seconds. This is due to those Pareto points coming from early, un-converged epochs.  

To be sure, there is significant nuance in differentiating ECGs. To again highlight CIS's discerning patience, we investigate the distribution of stopping times for each diagnosis compared to LARM. Figure \ref{fig:ecgPareto} (right) shows just this using CIS and LARM at 68\% accuracy (red circles in Figure \ref{fig:ecgPareto} (left)). We can see that CIS (i) on average stops sooner for NORM and MI diagnoses and (ii) has smaller interquartile ranges for all diagnoses. 

\subsection{Stock Option Experiment}
\begin{figure}[h!]
\centerline{
\includegraphics[width=0.5\linewidth, trim=0.6cm 0.4cm 2cm 2.2cm, clip]{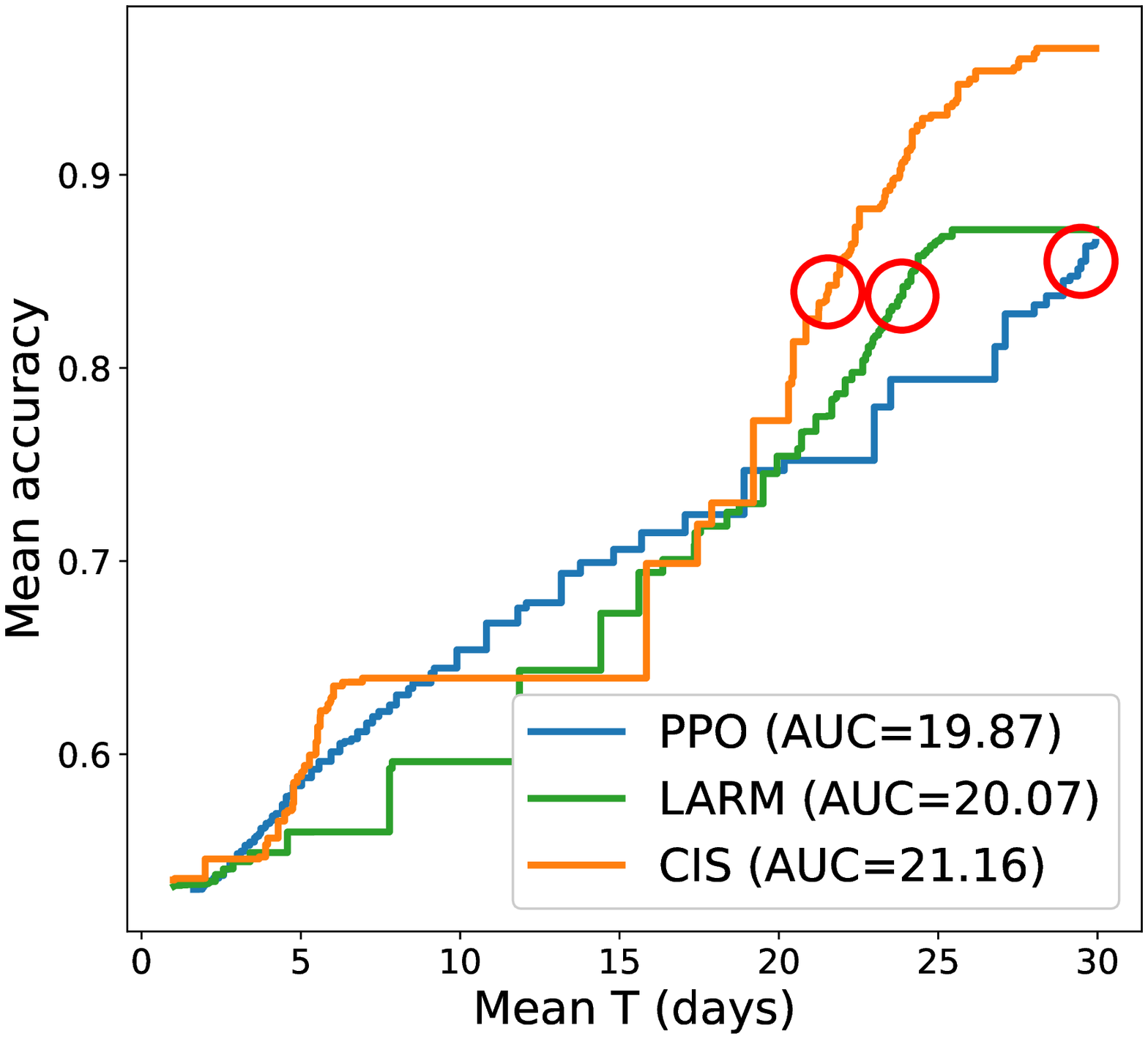} 
\includegraphics[width=0.5\linewidth, trim=0.6cm 0.4cm 2cm 2.2cm, clip]{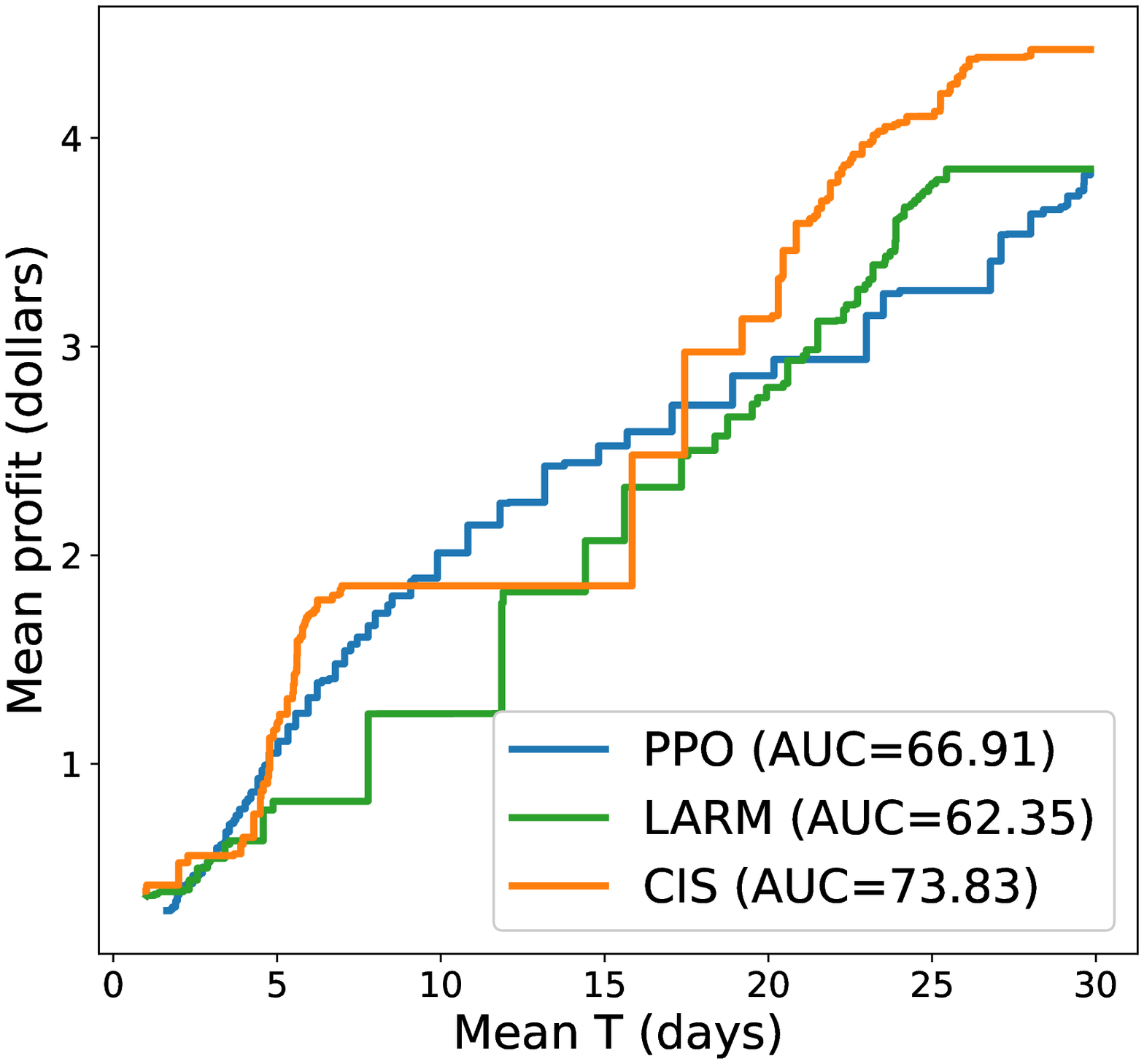} 
}
\caption{Accuracy (left) and profit (right) Pareto frontiers for the options experiment.}
\label{fig:stockPareto}
\end{figure}

For our stock option experiment, we construct two Pareto frontier comparisons, shown in Figure \ref{fig:stockPareto}. On the left, are the standard accuracy-time Pareto frontiers. However, in the financial scenario inspiring this experiment, dollars and profit is a more apt axis. So on the right, we also present profit-classification time Pareto frontiers. Here, we take a perfect hindsight definition of profit to include potential money gained and lost by not exercising the option. Since in this experiment we roll out the early classifiers continuously, the stochastic policies of PPO and LARM affect the future options (or samples). Accordingly, each Pareto frontier point is the average of 100 trials. One hundred trials is sufficient as the maximum ratio of standard error to mean is 2.7\% across all points' mean accuracies, profits, and classification times.

Again, our CIS holistically outperforms the benchmarks. In the accuracy sense, CIS's AUC is 6.5\% greater than PPO's AUC and 5.4\% greater than LARM's AUC. Turning to profit, CIS's AUC is 10.3\% greater than PPO's AUC and 18.4\% greater than LARM's AUC. 

\begin{figure}[h!]
\centerline{
\includegraphics[width=0.5\linewidth]{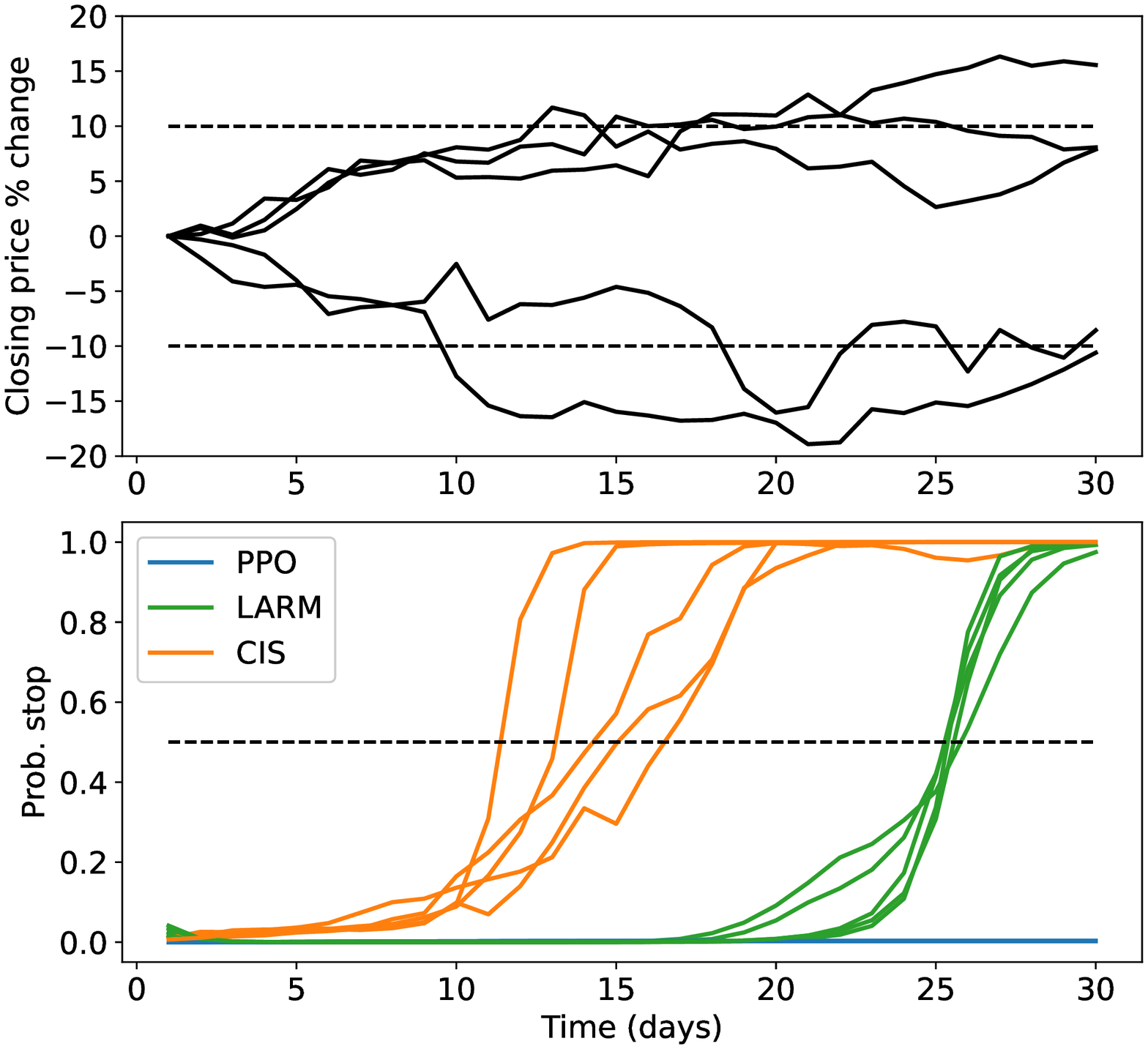} 
\includegraphics[width=0.5\linewidth]{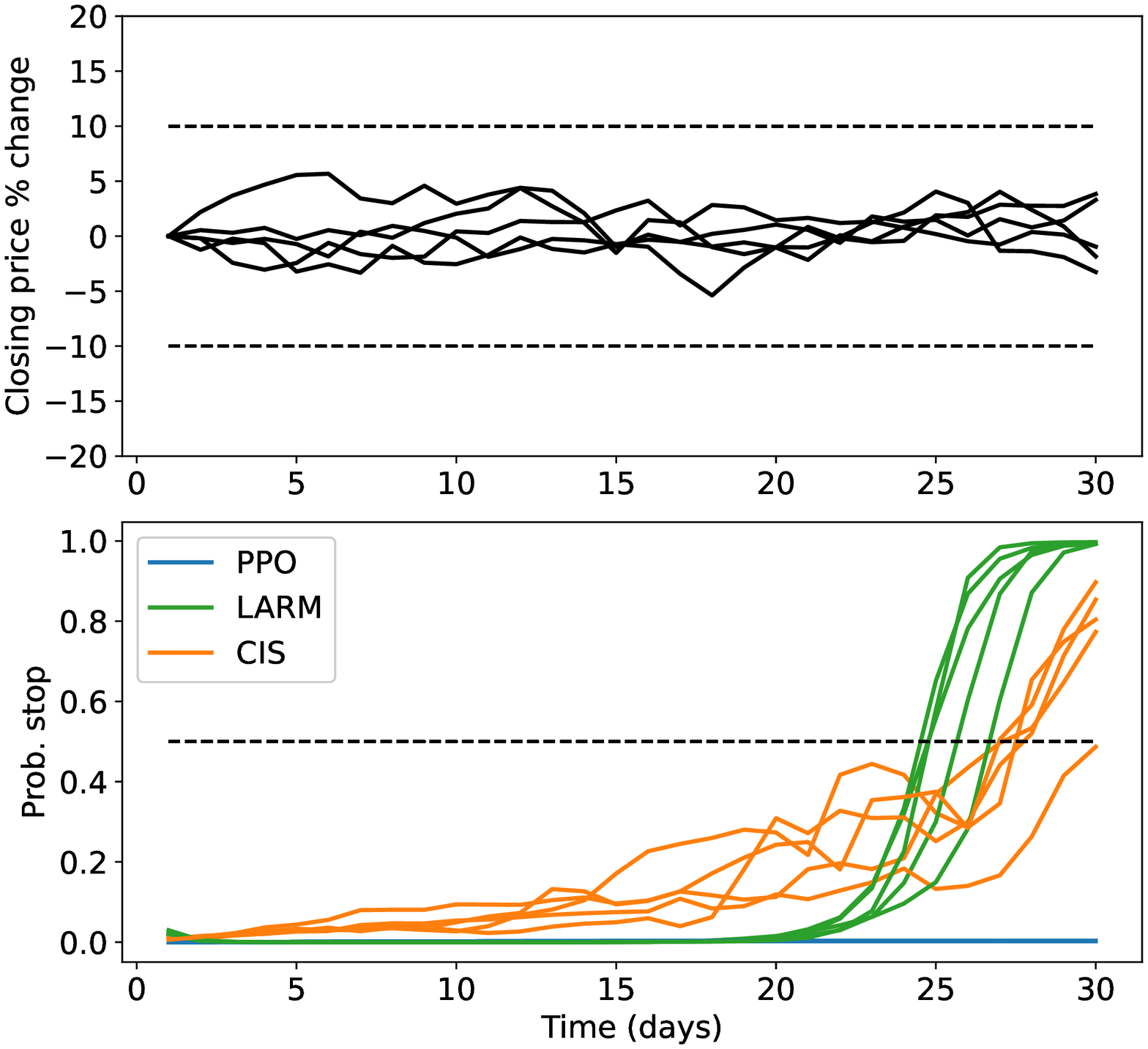} 
}
\caption{(Left plots) Five options with their daily percent change in closing price compared to day 1 exceeding $\pm 10\%$, marked with black dashed lines in the upper panel. We also mark 0.5 in the lower panel. (Right plots) Similar to left, except note the percent changes within $\pm 10\%$ for these options.}
\label{fig:stockExamples}
\end{figure}

While stock price movements are complex random walks, CIS is able to discern recognizable patterns better than LARM and PPO. If a stock displays strong and consistent growth or loss in the early days (drift in a random walk), one is more likely able to extrapolate a trend sooner. Similarly, if a stock continuously fluctuates around the first day's price, waiting longer becomes necessary to observe a trend, if any exist. Figure \ref{fig:stockExamples} demonstrates this hypothesis on an individual sample level using each early classifier at around 84\% mean accuracy (red circles in Figure \ref{fig:stockPareto} (left)). CIS stops much sooner for stock price movements with strong positive or negative drift. LARM consistently classifies around 25 days and PPO waits until the end (very low time penalty).

\section{Conclusion} \label{sec:conclude}
From our experiments, we stress CIS performs holistically better than state-of-the-art PPO and LARM in terms of a Pareto frontier AUC measure. On average, CIS is 3.6\% more accurate than LARM, and 19.8\% more accurate than PPO, given the same stopping time. Directly learning when to stop from its own classifications provides a better framework than exploration.

%
%
%
 \bibliographystyle{splncs04}
 \bibliography{unimodalEarlyClassifyICANN23}

\end{document}